\documentclass[conference]{IEEEtran}

\usepackage {multirow}

\usepackage{graphics}
 \usepackage{graphicx}
\usepackage{amssymb}

\usepackage{amsthm}

\usepackage{mdwmath}
\usepackage{mdwtab}
\usepackage{eqparbox}
\usepackage[cmex10]{amsmath}
\ifCLASSINFOpdf
\else
\fi

\begin{document}
%
\title{Offline Signature-Based Fuzzy Vault (OSFV):\\  Review and New Results}


\author{\IEEEauthorblockN{George~S.~Eskander, Robert~Sabourin and Eric~Granger}
\IEEEauthorblockA{\'Ecole de technologie sup\'erieure,  Universit\'e du Qu\'ebec\\
Montr\'eal, Canada\\
Email: geskander@livia.etsmtl.ca, robert.sabourin@etsmtl.ca, eric.granger@etsmtl.ca.\\ \\This paper has been submitted to \\The 2014 IEEE Symposium on Computational Intelligence in Biometrics and Identity Management (CIBIM)\\}}


%


\maketitle

\begin{abstract}
An offline signature-based fuzzy vault (OSFV) is  a bio-cryptographic implementation that uses handwritten signature images as biometrics instead of traditional passwords to secure private cryptographic keys. Having a reliable OSFV implementation is the first step towards automating financial and legal authentication processes, as it provides greater security of confidential documents by means of the embedded handwritten signatures. The authors have recently proposed the first OSFV implementation which is reviewed in this paper. In this system, a machine learning approach based on the dissimilarity representation concept is employed to select a reliable feature representation adapted for the fuzzy vault scheme. Some variants of this system are proposed for enhanced accuracy and security. In particular, a new method that adapts  user key  size  is presented. Performance of proposed methods are  compared using  the Brazilian PUCPR and GPDS signature databases and  results indicate that the key-size adaptation method  achieves a good compromise between security and accuracy. While  average system entropy is increased from 45-bits to about 51-bits, the AER (average error rate) is decreased by about 21\%.

\end{abstract}

%
\IEEEpeerreviewmaketitle

\section{Introduction}
Automation of financial and legal processes requires enforcement of confidentiality and integrity of transactions. For practical integration with the existing manual systems, such enforcement should be transparent to users. 
For instance, a person continually signs paper-based documents (e.g., bank checks) by hand, while his embedded handwritten signature images are used to secure the digitized version of the signed documents. 

 Such scenario can be realizable using  biometric cryptosystems (also known as bio-cryptographic systems \cite{Uludag2004}) by means of the offline handwritten signature images. In bio-cryptography,  biometric signals like fingerprints, iris, face or signature images, etc., secure private keys within  cryptography schemes like digital signatures and 
 encryption. Biometric samples provide a more trusted identification tool when compared to simple passwords. For instance, a fingerprint is attached to a person and it is harder to impersonate than traditional passwords.
 
 Despite its identification power, biometrics forms a  challenging design problem due to its fuzzy nature. For instance, while it is easy for a person to replicate his password  during authentication, it rarely happens that a person applies exact
 fingerprint each time.  The main source of variability in physiological biometrics like fingerprint, face, iris, retina, etc. is the imperfect acquisition of the traits. On the other hand, behavioral biometrics like handwritten signatures, gait, and even voice, have  intrinsic variability that is harder to cancel.
 
 Fuzzy vault (FV) is a reliable  scheme presented mainly to enable usage of fuzzy keys for cryptography  \cite{Jules2002}. 
 A FV decoder permits limited variations in the decryption key so that secrets can be decrypted even with variable keys.
 Accordingly, this scheme fits the bio-cryptography implementations, where biometrics are considered as fuzzy keys by which private cryptographic keys are secured.  Since the FV scheme has been proposed, it has being extensively employed for bio-cryptography, where  most implementations focused on physiological biometrics, e.g., fingerprints \cite{Nandakumar2007}, face \cite{Wang2007} and iris \cite{Lee2008}.   FV implementations based on the behavioral handwritten signatures are few and mostly employed online signature traits, where dynamic features like pressure and speed are acquired in real time by means of special devices as electronic pens and tablets \cite{Freire2006}. Static offline signature images, that are scanned after the signing process ends, however,  integrate too much variability to cancel by a FV decoder \cite{Freire2007}.

 Recently, the authors have proposed the first offline signature-based fuzzy vault  (OSFV) implementation \cite{Eskander2011}-\cite{Eskander2014-ICFHR}. This implementation is employed to design a practical digital signature system by means of handwritten signatures \cite{Eskander2013-wipra}. In this paper,   this implementation is reviewed and extended. In particular, we propose an extension to enhance the security and accuracy of the basic OSFV system by adapting cryptographic key size for individual users.   Finally,   system performance on the GPDS public signature database \cite{Vargas2007}, besides the private PUCPR Brazilian database \cite{Freitas2000}, are presented and interpreted.
 
 The rest of the paper is organized as follows. In the next section, the OSFV implementation  and its application
 to produce digital signatures by means of the handwritten signature images are reviewed. Section III describes the   signature representation and lists some aspects for enhanced representations.  Section IV introduces some OSFV variants for enhanced accuracy. Section V lists some variants for enhanced security. The new variant that adapts key sizes for enhanced security and accuracy is described  in Section VI. The simulation results are presented in Section VII.  Finally, some research directions and conclusions are discussed in Section VIII.

\section{Fuzzy Vaults With Signature Images}

\begin{figure*}[t]
 \begin{center}
 \includegraphics[scale=0.28]{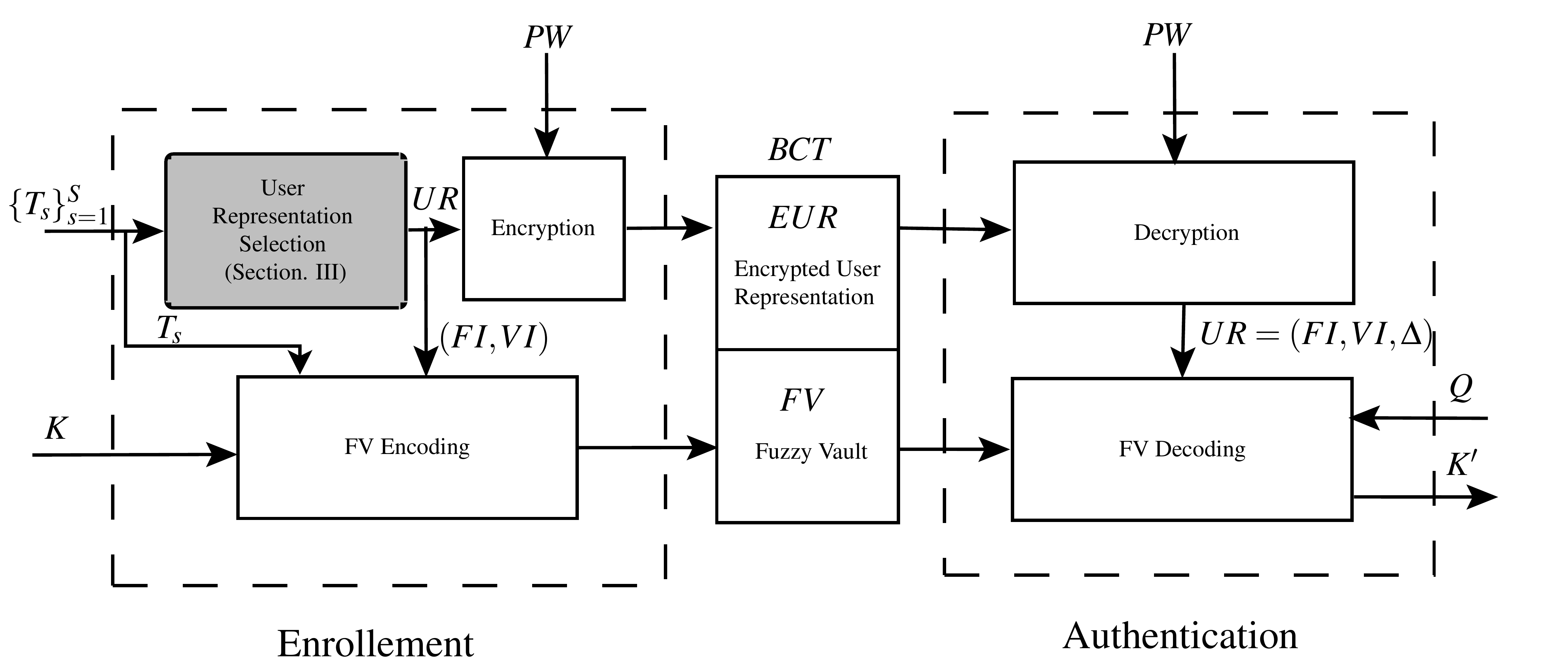}
\end{center}
\caption{Block diagram  of the  OSFV implementation \cite{Eskander2014-INS}.}
\label{fig:Figure6}
\end{figure*}

\begin{figure*}[t]

\begin{center}
 \includegraphics[scale=0.35,keepaspectratio=true]{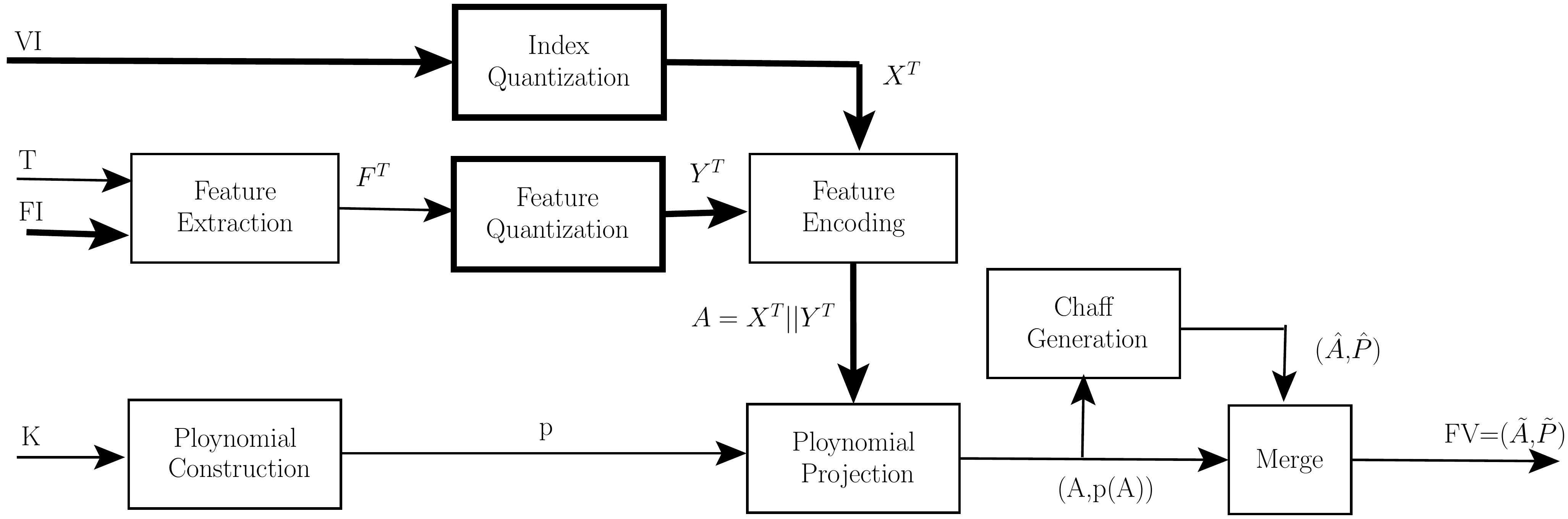}
\end{center}
 \caption{Block diagram of the  OSFV encoding process \cite{Eskander2014-INS}. The bold lines highlight the modules where specific modifications apply to the standard encoding process employed in the literature of biometric FVs, e.g., \cite{Nandakumar2007}.}
\label{fig:Figure7}
\end{figure*}

The system proposed for OSFV consists of two main sub-systems: enrollment and authentication  (see Figure \ref{fig:Figure6}).  In the enrollment phase, some signature templates $\{T_s\}_{s=1}^{S}$  are collected from the enrolling user. These templates are used for the user representation selection, as described in Section III.  The user representation selection process results in a user representations matrix $UR= (FI,VI,\Delta)$, where $FI=\{fI_i\}_{i=1}^{t}$ is the vector of indexes of the selected features,  $VI=\{vI_i\}_{i=1}^{t}$ is a vector of indexes mapping represented in $l/2$-bits\footnotemark[1]{} \footnotetext[1]{As will be described later in this section,   a feature index and its value are quantized in  $l/2$-bits each and then they are concatenated to produce a FV element of $l-bits$ size.  Since a feature index $fi$ might not fit in $l/2$-bits, we map it in a virtual index $v_i$ of a condensed size.}, and $\Delta=\{\delta_i\}_{i=1}^{t}$ is the vector of expected variabilities associated with the selected features.  This  matrix is user specific and contains important information needed for the authentication phase. Accordingly,  $UR$ is encrypted by means of a user password $PW$.  Both FV and password are then stored as a part of user bio-cryptography template ($BCT$).  Then, the user parameters $FI$ and $VI$ are used to lock the user cryptography key $K$  by means of  a single signature template $T_s$ in a fuzzy vault $FV$. 

In the authentication phase, user password $PW$ is used to decrypt the matrix $UR$. Then, the vectors $FI,VI$ and $\Delta$ are used to decode the FV by means of user query signature sample $Q$. Finally, user cryptographic key $K$ is released to the user so he can use it to decrypt some confidential information or digitally signs some documents.

\subsection{Enrollment process}

\begin{figure*}[t]
 \begin{center}
 \includegraphics[scale=0.34]{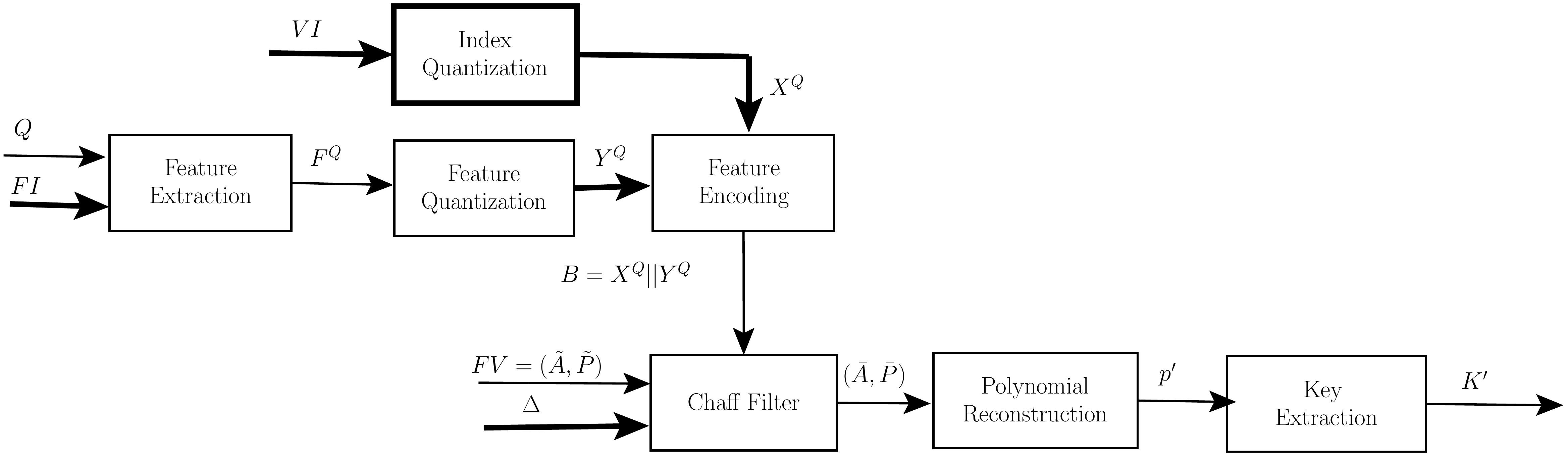}
\end{center}
\caption{Block diagram of the OSFV decoding process \cite{Eskander2014-INS}.  The bold lines highlight the modules where specific modifications apply to the standard encoding process employed in the literature of biometric FVs, e.g., \cite{Nandakumar2007}.}
\label{fig:Figure9}
\end{figure*}

The enrollment sub-system uses the user templates $\{T_s\}_{s=1}^{S}$, the password $PW$, and the cryptography key $K$ to generate a bio-cryptography template (BCT) that consists of the fuzzy vault $FV$ and the encrypted user representation matrix $EUR$. The user representation selection module  generates the $UR$ matrix as described in Section III.

The OSFV encoding module (illustrated in Figure \ref{fig:Figure7}) describes the following processing steps:

\begin{enumerate}
 \item the virtual indexes $VI=\{vI_i\}_{i=1}^{t}$ are quantized in $l/2$-bits and produces a vector $X^T=\{x^T_i\}_{i=1}^{t}$.
\item the user feature indexes $FI=\{f_i\}_{i=1}^{t}$ are used to extract feature representation $F^T=\{f^T_i\}_{i=1}^{t}$ from the signature template $T_s$. This representation is then quantized in $l/2$-bits and produces a vector $Y^T=\{y^T_i\}_{i=1}^{t}$.
\item The features are encoded to produce the locking set $A=\{a_i\}_{i=1}^{t}$, where $A=X^T||Y^T$ consists of $l$-bits FV points \footnotemark[2]{.}  \footnotetext[2]{FV points are represented with  fixed quantization sizes since FV decoders rely on error-correction codes that employ finite (Galois) field  computations.}
\item the cryptography key $K$ of size $KS$ where:

\begin{equation}
\label{KS}
 KS=l(k+1)-bits
\end{equation} 

  is split into $k+1$ parts of $l$-bits each \footnotemark[3]{}  \footnotetext[3]{The  FV quantization size $l$ is set to 16-bits in this work. So cryptographic keys of size $KS=128$-bits are encoded suing polynomials of degree $k=7$}, that constitutes a coefficient vector $C=\{c_0,c_1,c_2,....,c_k\}$. A polynomial $p$ of degree $k$ is encoded using $C$, where $p(x)=c_{k}x^k+c_{k-1} c^{k-1}+.....+c_{1} x+c_0$.
\item the polynomial is evaluated for all points in $A=\{a_i\}_{i=1}^{t}$ and constitutes the set $p(A)=\{p(a_i)\}_{i=1}^{t}$ where $p(a_i)=c_{k}{a_i}^k+c_{k-1} {a_i}^{k-1}+.....+c_{1} {a_i}+c_0$.

\item chaff (noise) points  $(\hat{A}=\{\hat{a_{ii}}\}_{{ii}=t+1}^{r},\hat{P}=\{\hat{p_{ii}})\}_{{ii}=t+1}^{r})$ are generated, where
$\hat{a_{ii}} \in GF(2^{l}), \hat{a_{ii}} \neq a_i \vee ii \in [t+1,r], i \in [1,t] $, and $\hat{p_{ii}} \in GF(2^{l}), \hat{p_{ii}} \neq p(\hat{a_{ii}}) \vee ii \in [t+1,r] $.  A chaff point $\hat{a_{ii}}=x_{ii}||y_{ii}$ is composed of two parts: the index part $x_{ii}$ and the value part $y_{ii}$. Two groups of chaff points are generated. Chaffs of $G_1$ have their indexes equal to the indexes of the genuine points. The chaff points and the genuine point that have the same index part are all equally spaced by a distance $\Omega$, eliminating the possibility to differentiate between the chaffs and the genuine point. Chaffs of $G_2$ have their index part differs than that of the genuine points \footnotemark[4]{.} \footnotetext[4]{The user password protects the $UR$ that stores his feature representation model.  If the attacker compromised the password, the indexes of the genuine points are known to him. In such case, chaffs of $G_
2$ are filtered out while $G_1$ could not be filtered without applying the good features.}
As the number of chaffs in $G_1$ is limited by the parameters $t$ and $\Omega$, so to inject higher quantity of chaffs we define $\alpha$ as a chaff groups ratio, where:

\begin{equation}
\label{alpha}
\alpha=g_2/g_1
\end{equation} 

where $g_1$ and $g_2$ are the amount of chaff features belong to $G_1$ and $G_2$, respectively. $G_2$ chaffs are generated with $\alpha t$  indexes different than the $t$ genuine indexes.  Hence, the FV size $r$ is given by:

\begin{equation}
\label{r}
r=t(\alpha+1)/\Omega
\end{equation} 

So, the total number of chaffs $z$ is given by:

\begin{equation}
\label{z}
z=t(\alpha+1-\Omega)/\Omega
\end{equation} 

\item  the genuine set  $(A,p(A))$, and the chaff set $(\hat{A},\hat{P})$ are merged to constitute the fuzzy vault $FV=(\tilde{A},\tilde{P})$, where $\tilde{A}= A  \bigcup \hat{A}$,  $A=\{a_i\}_{i=1}^{t}$, $ \hat{A}=\{\hat{a}_i\}_{i=t+1}^{r}$ and  $\tilde{P}= p(A)  \bigcup \hat{P}$, $p(A)=\{p(a_i)\}_{i=1}^{t}, \hat{P}=\{\hat{p}_i\}_{i=t+1}^{r}$.
\end{enumerate}

\subsection{Authentication Process}

The authentication sub-system uses the user query sample $Q$ and the password $PW$, to decode the fuzzy vault $FV$ and restore the user cryptography key $K$. First the password $PW$ is used to decrypt the $UR$ matrix. Then the vectors $FI,VI$, and $\Delta$ are used to decode the FV by means of the query $Q$. 
  
\begin{figure*}[t]
\begin{center}
 \includegraphics[scale=0.29]{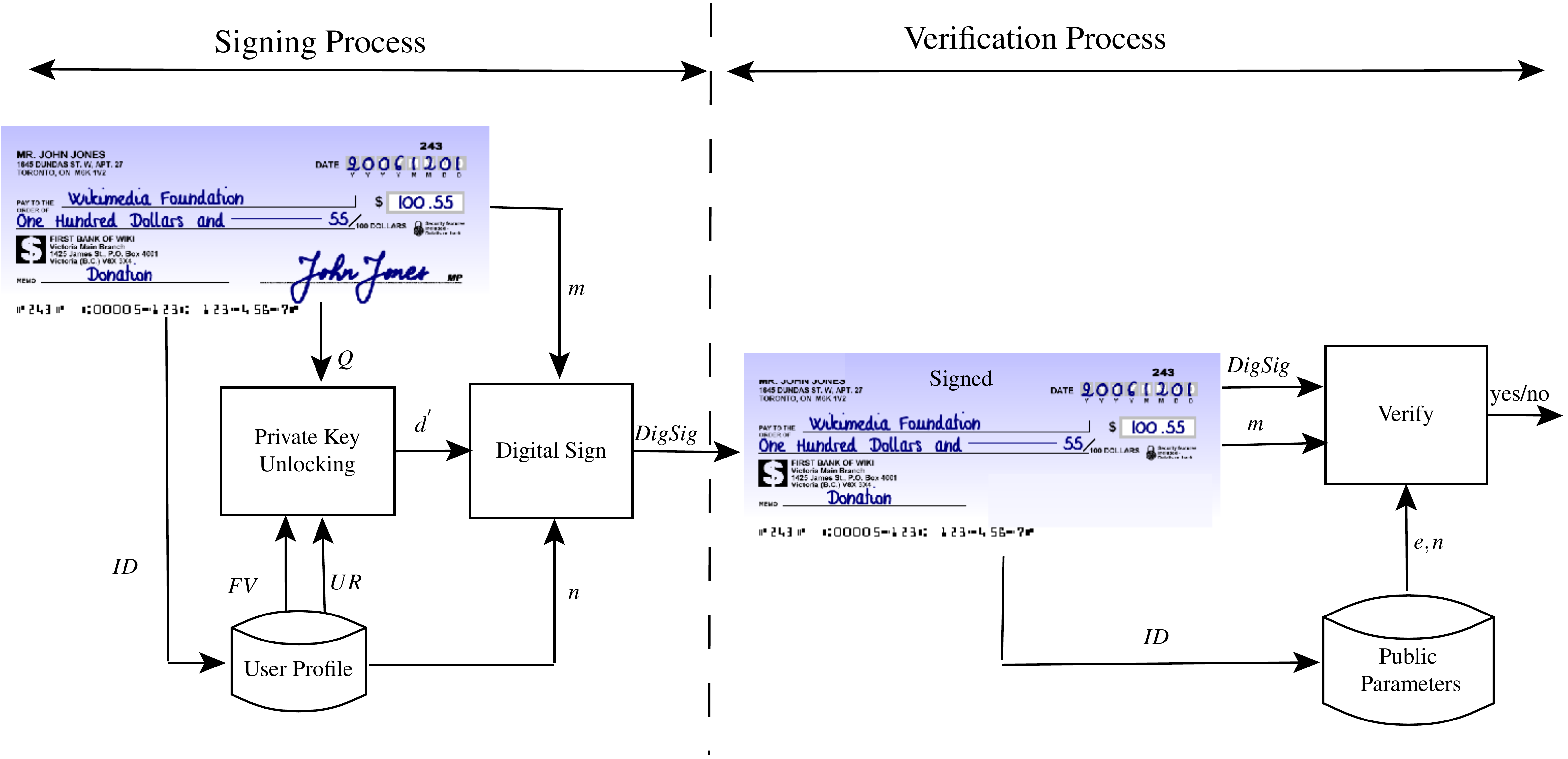}
\end{center}
\caption{Framework of a OSFV-based digital signature method \cite{Eskander2013-wipra}.} 

\label{signature}
\end{figure*}

The OSFV decoding module (illustrated in Figure \ref{fig:Figure9}) describes the following processing steps:

\begin{enumerate}
\item the virtual indexes $VI=\{vI_i\}_{i=1}^{t}$ are quantized in $l/2$-bits and produces a vector $X^Q=\{x^Q_i\}_{i=1}^{t}$.
 \item the user feature indexes $FI=\{f_i\}_{i=1}^{t}$ are used to extract feature representation $F^Q=\{f^Q_i\}_{i=1}^{t}$ from  $Q$. This representation is then quantized in $l/2$-bits and produces a vector $Y^Q=\{y^Q_i\}_{i=1}^{t}$.
\item The features are encoded to produce the unlocking set $B=\{b_i\}_{i=1}^{t}$, where $B=X^Q||Y^Q$. Hence, the unlocking elements are represented in a field $GF(2^{l})$.
\item the unlocking set $B$ is used to filter the chaff points from the FV. An adaptive matching method is applied to match unlocking and locking points. Items of $B$ are matched against all items in $\tilde{A}$.  This process results in a  matching set $(\bar{A},\bar{P})=((B \bigcap \tilde{A}), p\leftarrow (B \bigcap \tilde{A}))$, where
$p\leftarrow (B \bigcap \tilde{A})$ represents the projection of the matching features on the polynomial space. Chaff filtering is done as follows. If the feature indexes are correct \footnotemark[5]{} \footnotetext[5]{That occurs if the applied password is genuine, so the $UR$ is decrypted properly and the right indexes are restored.}, then  all elements of $X^Q$ will have corresponding elements in $X^T$. So, all of chaffs of $G_2$ will be filtered out. Then, each of the remaining FV points will be compared to corresponding points extracted from the query sample. An adaptive matching method is applied:  for every feature $i$, a matching window $w_i$ is adapted to the feature modeled variability $\delta _i$,  where $w_i=2\delta_i$. A FV point $a_i$ is considered matching with an unlocking point $b_i$, if they reside in the same matching window. I.e., $|a_i-b_i| \le w_i$. 

\item the matching set $(\bar{A},\bar{P})$ is used to reconstruct a polynomial $p'$ of degree $k$ by applying the RS decoding algorithm \cite{Berlekamp1968}.
\item the coefficients of $p'$ are assembled to constitute the secret cryptography key $K'$.
\end{enumerate}

\subsection{Applications}

 In \cite{Eskander2013-wipra}, the OSFV implementation is employed  to produce
digital signatures using offline handwritten signatures. This methodology facilitates the automation of business processes, where users continually employ their handwritten signatures for authentication. Users are
isolated from the details related to the generation of digital signatures,
yet benefit from enhanced security. 

Figure \ref{signature} illustrates the OSFV-based digital signature framework. The user FV that is constructed during enrollment is used to sign  user documents offline as follows.  
When a user signs a document by hand, his handwritten signature image is employed to unlock his private key $d^{'}$. The unlocked key produces
a digital signature by encrypting some message $m$ extracted from the document (e.g., check amount).  The encrypted message is considered as a digital signature and it is attached to the digitized document. Any party who possesses the user public key can verify the digital signature, where verification of the digital signature  implies authenticity of
the manuscript signature and integrity of the signed document (e.g., check amount did not change). 


\section{Feature Representation}

\begin{figure}[t]
	\centering
	\includegraphics[scale=0.19]{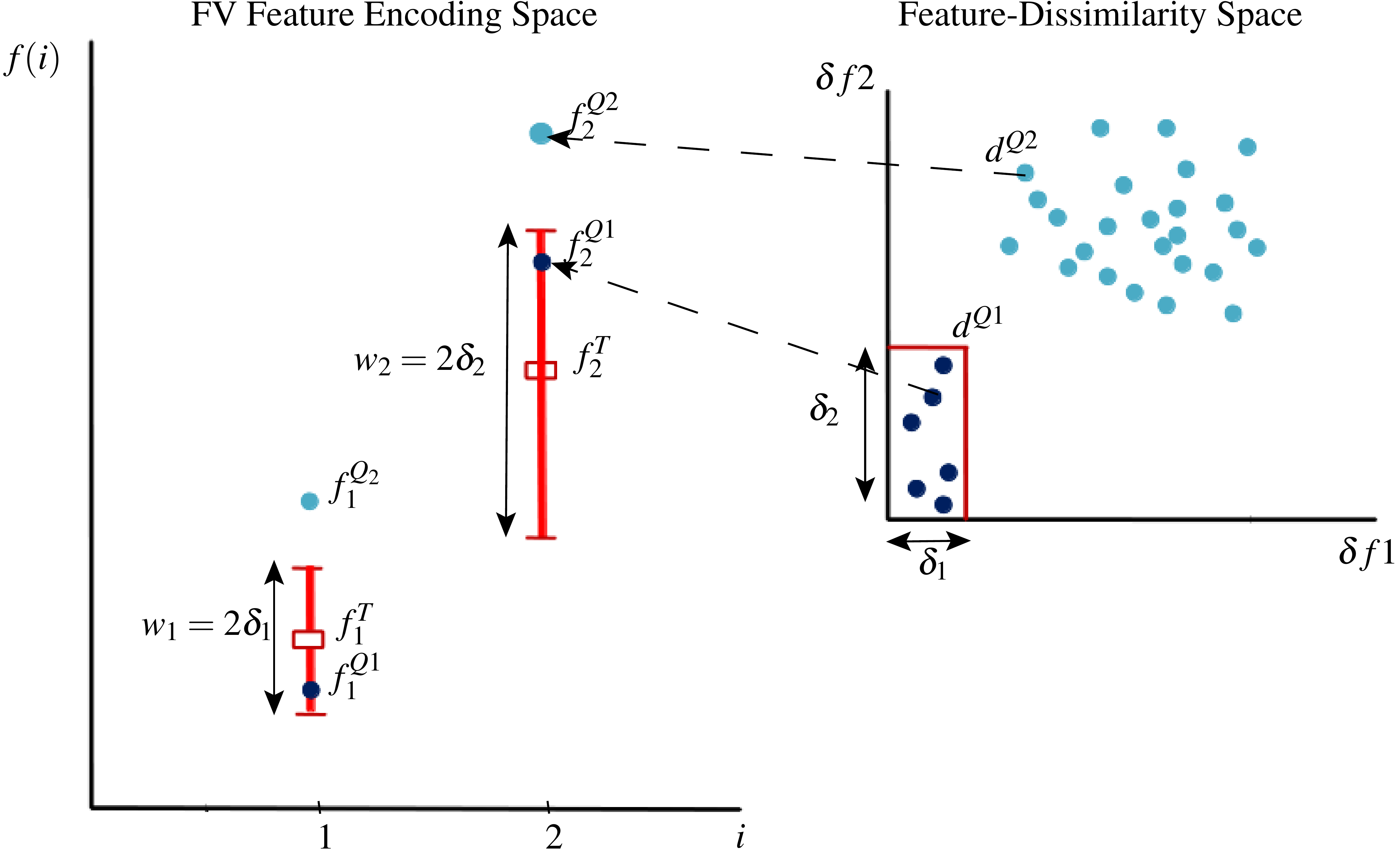}
	\caption{Illustration of the feature representation designing process \cite{Eskander2014-INS}.}
	\label{fig:Figure12}
\end{figure}

According to aforementioned OSFV implementation, the FV points encode some features extracted from the signature images. It is obvious that accuracy of a FV system relies on  the  feature representation. Representations of intra-personal signatures should sufficiently overlap so that matching errors lie within the error correction capacity of the FV decoder.  On contrary, representations of inter-personal signatures should sufficiently differ so that matching errors are higher than the error correction capacity of the FV decoder.

Accordingly, the authors proposed to design signature representations adapted for the FV scheme by applying a feature selection process in a feature-dissimilarity space.  In this space,   features are extracted from 
each pair of template and query samples and  the pair-wise feature distances are used as space dimensions. 

\begin{figure*}[t]
 \begin{center}
 \includegraphics[scale=0.28]{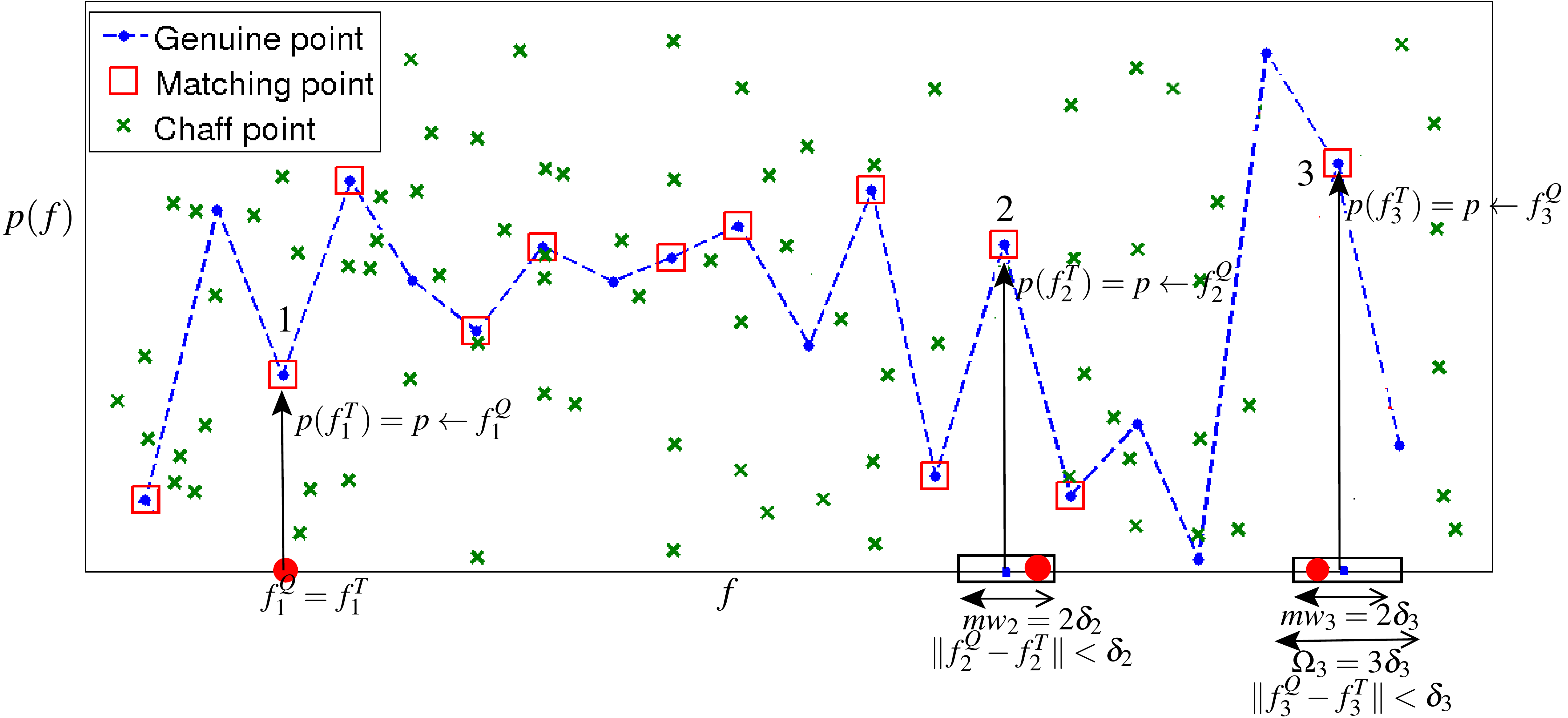}
\end{center}
\caption{Illustration of adaptive matching and adaptive chaff generation methods \cite{Eskander2013-EAHSP}.} 
\label{fig:Figure3}
\end{figure*}

To illustrate this approach, see Figure \ref{fig:Figure12}.  In this example, three signature images are represented: $T$ is the template signature, $Q_1$ is a genuine  query sample and $Q_2$ is a forgery query sample. In the left side, signatures are represented in the FV feature encoding space, where a FV point encodes a feature index $i$ and its value $f_i$. For simplicity, only two features ($f_1$ and $f_2$) are shown, while the full representation consists of $t$ dimensions.  On the right side, signatures are represented in the feature dissimilarity space. In this space, a feature is replaced by its distance from a reference value. For instance,  $f_1$ and $f_2$ are replaced by their dissimilarity representations $\delta f_1, \delta f_2$, where $\delta f_1= |f_1^Q-f_1^T|$, and  $\delta f_2= |f_2^Q-f_2^T|$. Accordingly, while a point in the feature encoding space represents a signature image, a point in the feature dissimilarity space represents the dissimilarity between two different signature images. The point $d^{Q_1}$ 
represents the 
dissimilarity between the genuine signature $Q_1$ and the template $T$, and  a point $d^{Q_2}$ represents the dissimilarity between the forgery signature $Q_2$ and the template $T$, where $d^{Q_1}$= ($\delta f_1^{Q_1}, \delta f_2^{Q_1},.....,\delta f_t^{Q_1}$), and $d^{Q_2}$= ($\delta f_1^{Q_2}, \delta f_2^{Q_2},.....,\delta f_t^{Q_2}$).

In this example, $\delta f_1$ and $\delta f_2$
are discriminant features. For instance,  for all genuine query samples like $Q_1$, $\delta f_i^{Q_1} < \delta_i$ and for all forgery query samples like $Q_2$, $\delta f_i^{Q_2} > \delta_i$. Unfolding these  discriminant  dissimilarity features to the original feature encoding space produces discriminant features in the encoding feature space, where the distance between two feature instances is used to determine their similarity. For instance, a genuine feature (like $f_i^{Q_1}$) lies close to the template feature $f_i^T$, so they are similar, where closeness here implies that both features reside in a matching window $w_i=2\delta_i$.  Features extracted from a forgery image (like $f_i^{Q_2}$) do not resemble the template feature $f_i^T$, as they reside outside the matching window $w_i$.

Aforementioned  description of the process to design  representations   is generic.  Some   extensions are reviewed and compared below.

\subsection{Global VS Local Representations}

Shortage of user  samples for training is addressed by designing a global writer-independent (WI) representation \cite{Eskander2011}. A large number of signature images from a development database  are represented in the feature-dissimilarity space of high dimensionality, and feature selection process runs to produce a global space of reduced dimensionality.  Such global approach permits designing FV systems for any user even who provides a single signature sample during enrollment. 

For performance improvement, the global representation is specified for individual users once enough number of enrolling samples becomes available. To this end, training samples are firstly represented in the global representation space, then an additional training step runs to produce  a local writer-dependent (WD) representation that discriminates the specific user from others. Simulation results have shown that local representations enhanced FV decoding accuracy by about 30\%, where the average error rate (AER) is decreased from 25\% in case of global representations to  17.75\% for the local ones.

\subsection{Multi-Scale Feature fusion}

In \cite{Eskander2011}, the Extended Shadow Code (ESC) feature extraction method is adapted for the FV implementation  \cite{Sabourin1994}. These features consist in the superposition of a bar mask array over a binary image of handwritten signatures. Each bar is assumed to be a light detector related to a spatially constrained area of the 2D signal.  This method is powerful in detecting various levels of details in the signature images by varying the extraction scale. For instance, an image could be
split to $h \times v$ of horizontal and vertical cells, respectively, and shadow codes are extracted within individual cells. The higher the number of cells, the higher the resolution of detectors.   

The authors observed that designing FVs based on a single extraction scale results in varying performance for the different users. For  instance, while high resolution scales are fine with users whose signatures are easy to forge or those who have high similarities with others, the low resolutions are better for users whose signatures integrate high variabilities. Accordingly, a multi-scale feature fusion method is proposed, where different feature vectors are extracted based on different extraction scales and they are combined to produce a high-dimensional representation. This representation is then processed through the WI and WD design phased and produces the final local representation that  encodes in the FV. 

\begin{figure*}[t]	
\begin{center}
\includegraphics[scale=0.28]{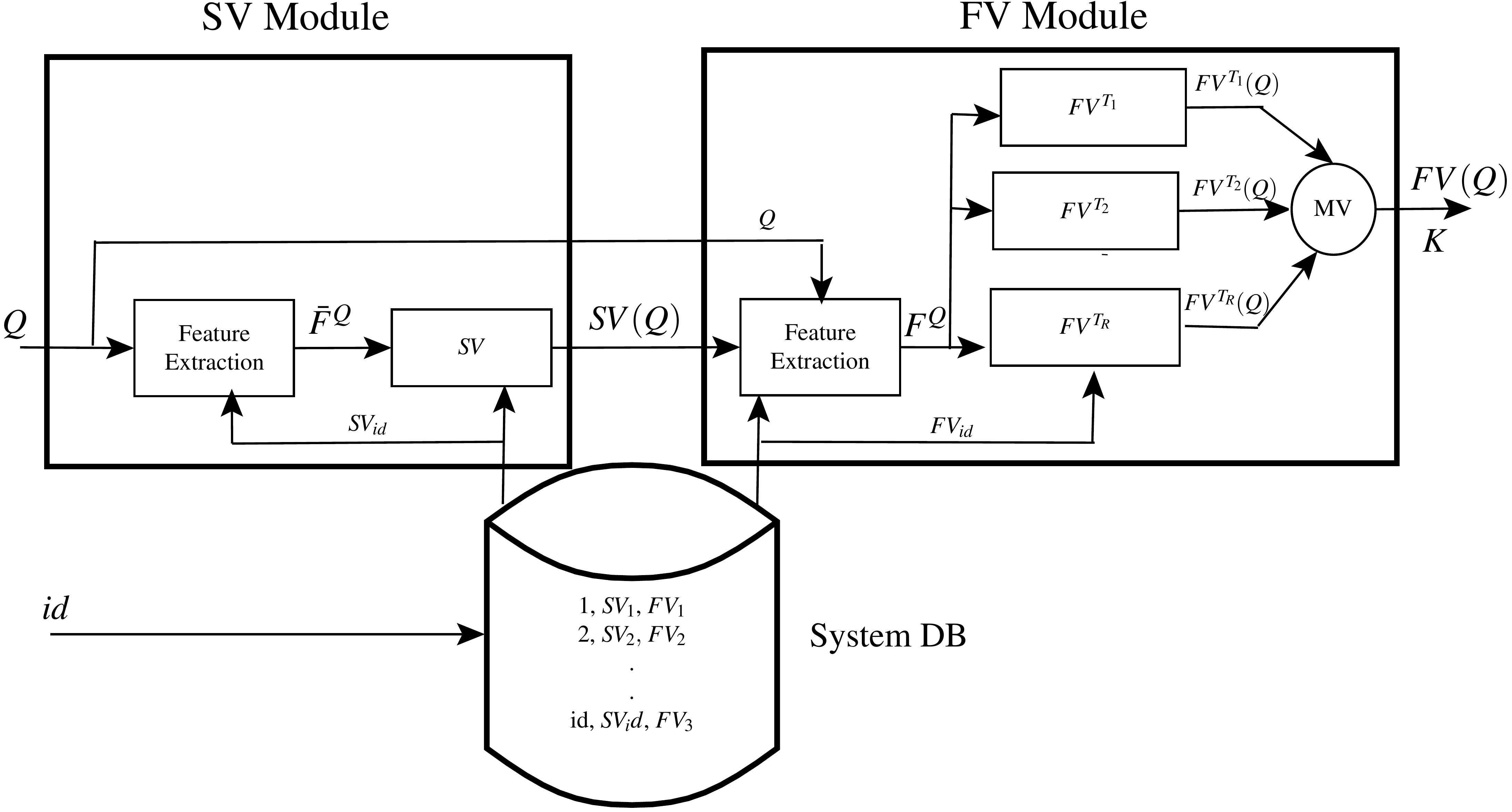}
\end{center}
\caption{Cascaded SV-FV system in the verification mode: different feature representations $\bar{F}$ and $F$ are processed by a SV classifier and a set of FV decoders, respectively. The FV module is triggered only if the SV module produces a positive classification label \cite{Eskander2014-ICFHR}.}
\label{SV-FV}
\end{figure*}

\subsection{Multi-Type Feature fusion}

Besides fusing feature vectors that are extracted based on different scales, it is possible to fuse different types of features.
In \cite{Eskander2013-EAHSP}, the directional probability density function (DPDF) features \cite{Drouhard1996} are fused with  ESC features to constitute a huge dimensional representation (of 30,201 dimensionality). This representation is reduced through the WI and WD training steps and produced a concise representation of only 20 features. It is shown that injecting the additional feature type increased the FV decoding accuracy by about 22\% (AER is reduced from 17.75\% to 13.75\%).

\subsection{Prototype Selection}

The aforementioned approach provides a practical scenario to produce representations with low intra-personal and high inter-personal variabilities which is mandatory feature for FV systems. However, the authors observed that the margin between the intra and the inter classes differs when using different signature prototypes (templates) for FV encoding. Accordingly, a prototype selection method is proposed  \cite{Eskander2013-SIMBAD}. The WD representation is projected  to a dissimilarity space where distances to different user prototypes are the space constituents. Then, a  feature selection process runs in the dissimilarity space and locates the best prototype.  This method has enlarged the separation between the intra and inter clusters significantly (Area under ROC curve (AUC) is increased from 0.93 to 0.97).

\section{Extensions For Enhanced Accuracy}

Although  accuracy of an OSFV system  relies mainly on quality of the  feature representation, the proposed implementation provides additional opportunities for  enhanced accuracy  by applying some other design  variants as described in this section.

\subsection{Adaptive Matching}

The results mentioned so far report accuracy of FV decoders that apply strict matching approach. Two FV points are matching only if they have identical values. Accuracy of a FV decoder is enhanced by applying the adaptive matching method, where the feature variability matrix $\Delta$ is used for matching so that corresponding FV points are considered matching if their difference lies within the expected variability of their encoding feature (see Figure \ref{fig:Figure3}). This method increased accuracy by about 27\%  (AER is reduced from 13.75\% to 10.08\%) \cite{Eskander2014-INS}.

\subsection{Ensemble of Fuzzy Vaults}

Instead of decoding a single FV token, it is possible to decode several FVs for enhanced performance. In case that 
some FVs are correctly decoded, the decrypted key is released to the user based on the majority vote rule. This method  has increased detection accuracy by about 18\%  (AER is reduced from 10.08\% to 8.21\%) \cite{Eskander2014-INS}.

\subsection{Additional Passwords}
The limited  discriminative power of  FVs is alleviated by using an additional password $PW$, so that the false accept rate (FAR) is reduced without significantly affecting the false reject rate (FRR). For the results reported so far, it was  assumed that the user password $PW$ is compromised. However, to report the actual
performance of the system we have to consider the case when an attacker neither possesses a correct password nor a genuine signature sample. In this case, he cannot decrypt the UR model and hence he randomly guesses the feature indexes.  It is shown that the additional password has increased detection accuracy by about 65\%  (AER is reduced from 8.21\% to 2.88\%) \cite{Eskander2014-INS}.

\subsection{Cascading With Traditional SV Modules}

Using additional passwords for  enhanced system accuracy comes with the expense of the user inconvenience. In \cite{Eskander2014-ICFHR}, a novel user-convenient approach is proposed for enhancing the accuracy of signature-based biometric cryptosystems. Since signature verification (SV) systems designed in the original feature space have demonstrated higher discriminative power to detect impostors \cite{Eskander2013IET}, they can be used to improve the FV systems. Instead of using an additional password, the same signature sample is processed by a  SV classifier before triggers the FV decoders (see Figure \ref{SV-FV}). Using  this cascaded approach, the high FAR of FV decoders is alleviated by the higher  capacity of SV classifiers to detect impostors. This method  has increased detection accuracy by about 35\%  (AER is reduced from 10.08\% to 6.55\%). When multiple FVs are fused, the AER is decreased by 31.30\% (from 8.21\% to 5.64\%).

\section{Extensions For Enhanced Security}
Security of the  OSFV implementation is analyzed in terms of the brute-force attack \cite{Eskander2014-INS}. Assume an attacker could compromise the FV without possessing neither valid password nor genuine signature sample. In this case, the attacker tries to separate enough number of genuine points ($k+1$) from the chaff points. Security of  a FV is given by:

\begin{equation}
\label{s}
security \cong \binom{(\alpha+1)t}{k+1} (1/ \Omega)^{k+1}
\end{equation} 

Where $\alpha$ is the chaff group ratio, $t$ is the number of genuine points in the FV, $k$ is the degree of the encoding polynomial and $\Omega$ is the chaff separation distance.

High value of $\alpha$ implies a high number of G2 features which are compromised in case  that the password is compromised.  The parameter $t$ should be concise as it impacts the accuracy and complexity of the FV. Accordingly,  entropy of the system can be increased through using different values of the parameters: $\Omega$ and $k$.  However,  there is a trade-off between system security and its recognition accuracy that could be alleviated by applying the following approaches.

\subsection{Adaptive Chaff Generation}

In the traditional chaff generation method,  equal-spaced chaff points are generated with a separation factor $\Omega$ \cite{Eskander2014-INS}. In such case, there is a trade-off between security and
robustness. For instance, with small separation, e.g., $\Omega=0.025$, there are 40 FV points
generated with the same index (1 genuine + 39 chaff points). In this case, a high
number of chaffs is generated and results in high system entropy of about 68-bits and low accuracy  of about  $20\%$ AER.

The adaptive chaff generation method enables the injection of high number of chaff with minimal impact on the FV decoding robustness. To this end,  the feature variability vector
$\Delta$ is used during the FV locking phase so that chaff points
are generated adaptively according to feature variability.  For each feature $f_i$ ,  $\Omega_i = 3 \times \delta_i$  (see Figure \ref{fig:Figure3}). By this method, it is less likely that an unlocking element equates a chaff element. For or instance, the same entropy (68-bits) could be achieved  with a minimal impact on  system robustness
(AER = 10.52\%)  \cite{Eskander2013-EAHSP}.

\subsection{Controlling Key Size}

According to Eq.\ref{s}, the longer the cryptographic key size the higher entropy of the FV. However, this comes with expense 
of the accuracy \cite{Eskander2014-INS}. In \cite{Eskander2013-wipra}, different key sizes (KS) are tried (128, 256, 512, 1024-bits) and it is shown that  different key sizes result in different performance for the
different users. This observation motivates  adapting the key length
for each user as proposed in the following section,

\section{The Adaptive Key Size Approach}

In \cite{Eskander2013-EAHSP}, functionality of a FV decoder is formulated as a simple dissimilarity threshold as follows:

{\baselineskip=-1 pt
\begin{center}
\begin{equation}
\label{FVd}
FV^{T}(Q)=sign(\epsilon-(\delta_A^{QT}+\delta^{'})).
\end{equation}
\end{center}
}

Where a FV encoded by a template $T$ can be correctly decoded by a query $Q$ only if the total dissimilarity between $Q$ and $T$  is less than the error correction capacity $\epsilon$ of the FV decoder. Here, $\delta_A^{QT}$ is the dissimilarity part that results from the variability between the two samples, and $\delta^{'}$ is the dissimilarity part that results from wrong matches with chaff points. 

The methods discussed so far aimed to optimize the dissimilarity parts of Eq.\ref{FVd}. For instance, the multi-scale and multi-type feature extraction approach results in separating intra-personal and inter-personal dissimilarity ranges. Selection of robust templates (prototypes) and applying adaptive matching enlarged this separation. Also, impact of the chaff error is minimized by presenting the adaptive chaff method. With applying all these methods, however, accuracy of a signature-based FV is still below the level required for practical applications. Accordingly, performance is increased by applying some complex and user inconvenient solutions like ensemble of FVs and using additional passwords or cascading SV and FV systems. 

Here we investigate a new room for enhancing FVs by optimizing the error correction capacity $\varepsilon$ which is given by:

{\baselineskip=-1 pt

\begin{center}
\begin{equation}
\label{FV}
 \varepsilon=(t-k-1)/2
\end{equation}
\end{center}
}

It is obvious that this parameter relies on the FV encoding size $t$ and the encoding polynomial degree $k$. Also, from Eq.\ref{KS}, we see that $k$ determines the key size $KS$. Accordingly, we select user specific key sizes through changing the parameter $k$ so that $\varepsilon$ for a specific user covers the range of his expected signature variability. To this end, we set $\varepsilon$ for a user to his  maximum intra-personal variability $e$.
Based on the resulting user-specific error correction capacity, the parameter $k$ is determined using Eq.\ref{FV} and user key size $KS$ is computed using Eq.\ref{KS}\footnotemark[6]{} \footnotetext[6]{ In this work, we set an upper limit for the error correction capacity to be  $\varepsilon \le 6$ so that $k \ge 7$ and $KS \ge 128$-bits.}.

Once appropriate key size is computed for a user, his key is enlarged through injecting some padding bits in the
original key during FV encoding. During authentication, the enlarged key is reconstructed and the padding bits are removed to produce the original cryptographic key.

\section{Simulation Results}

All aforementioned performance results are reported for the PUCPR Brazilian signature database \cite{Freitas2000}.
Here we test the system for the public GPDS-300 database \cite{Vargas2007} as well. This database contains signatures of $300$ users, that were digitized as 8-bit greyscale at resolution of 300 dpi and contains images of different sizes (that vary from $51 \times 82$ pixels to $402 \times 649$ pixels). 
All users have 24 genuine signatures and 30 simulated forgeries. It is split into
two parts. The first part contains signatures of the first 160  users. A subset of this part is used to design the local representation  and the remaining of this part is used for  performance evaluation. The second part contains signatures of the last 140 users and it is used to design the global representation. See \cite{Eskander2013IET} for a similar experimental protocol for both databases.

Table \ref{table:Impact of Using a User Password} shows results for the two databases for fixed and adaptive key sizes. It is obvious that employing the adaptive key size approach decreased the FAR significantly with low impact on the FRR. For instance,
the AER for the PUCPR database in decreased by about 21\% (from 10.08 to 7.94). Also, performance of the system for the GPDS database is comparable to state-of-the-art traditional SV systems (AER is about 15\%) that employ more complex classifiers  \cite{Eskander2013IET}.

\begin{figure}
\begin{center}

 \includegraphics[scale=0.30]{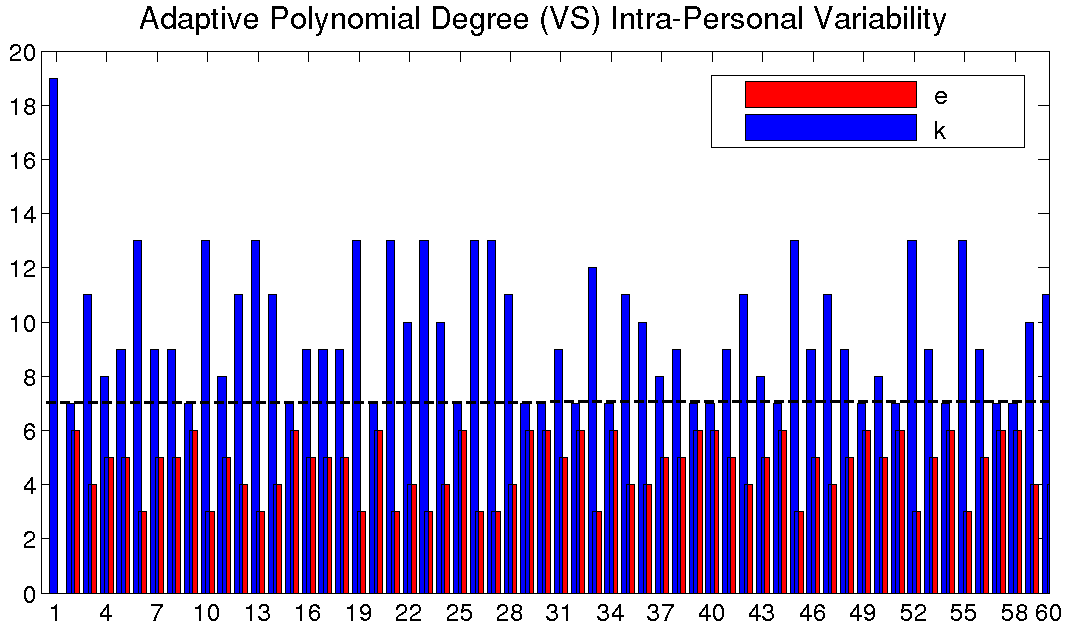}
 \caption{Adaptive polynomial degree (k) VS Intra-personal variability (e)}
 \label{fig:adaptive}
 \end{center}
 
\end{figure}

Moreover, the proposed method also enhances system security as it is possible to increase the key size, and hence the polynomial degree $k$, without much impact on the accuracy. For instance, Figure \ref{fig:adaptive} shows the adapted polynomial degrees for different users in the PUCPR database and the corresponding user variability $e$. It is obvious that users with more stable signatures have their cryptographic keys more enlarged than users with less stable signatures. According to Eq.\ref{s}, system entropy of the standard OSFV implementation (with fixed keys of size 128-bits and polynomial degree $k=7$) is about 45-bits. With applying the adaptive key size method, the  average $k$ is about 9.6-bits (see Figure \ref{fig:adaptive}) which provides an average entropy of about 51-bits.

\begin{table}[t]
\caption{Impact of Using a User Password as a second Authentication Measure}
\label{table:Impact of Using a User Password} 
\centering 
\begin{tabular}{|c|c|c|c|c|} 

\hline\multirow{2}*{Measure}&\multicolumn{2}{|c|}{PUCPR DB}& \multicolumn{2}{|c|}{GPDS DB}\\
\cline{2-5}&Fixed Key& Adaptive Key&Fixed Key&Adaptive Key\\
\hline
$FRR$&11.53&12.71&37.5&39.07\\
\hline 
$FAR_{random}$&2.05&0&0&0\\
\hline 
$FAR_{simple}$&2.39&0.05&-&-\\
\hline 
$FAR_{simulated}$&24.28&19.02&15.37&11.20\\
\hline 
$AER_{all}$&10.08&7.94&17.6&16.75\\
\hline 
\end{tabular}
\end{table}

\section{Conclusions and Research Directions }

In this paper, a recently published offline signature-based FV implementation is reviewed. Several variants of the system are listed and compared for enhanced accuracy and security. A novel method to adapt cryptography key sizes for different users is proposed and have shown accuracy and security enhancement. The performance is also validated on a public signature database where comparable results of complex SV in the literature is reported. Although the proposed key adaptation method sounds, there is need to propose more intelligent tuning technique taking in consideration the similarities with simulated forgeries for higher forgery detection. This study listed many new approaches that are applied successfully to the signature based bio-cryptography. We believe that these methods shall be investigated for other biometrics which might enhance state-of-the-art of the area of  bio-cryptosystems.






%

\end{document}